\documentclass[twocolumn]{article}
\usepackage{amsmath,epsfig}

\usepackage{amsmath}
\usepackage{bm}
\usepackage{url}
\usepackage{svg}
\usepackage{tikz, graphicx}
\usepackage{standalone}
\usepackage{hyperref}
\usepackage[T1]{fontenc}
\usepackage[normalem]{ulem}

\usepackage{caption}
\newcommand\scalemath[2]{\scalebox{#1}{\mbox{\ensuremath{\displaystyle #2}}}}

\title{Automatic Stockpile Volume Monitoring \\ using Multi-view Stereo from SkySat Imagery}

\author{Roger Mar\'{i}$^1$ \hspace{0.7 cm} Carlo de Franchis$^{1,2}$ \hspace{0.7 cm} Enric Meinhardt-Llopis$^1$ \hspace{0.7 cm} Gabriele Facciolo$^1$}
\date{$^1$Universit\'{e} Paris-Saclay, CNRS, ENS Paris-Saclay, Centre Borelli, France \\ $^2$Kayrros SAS}

\begin{document}

\maketitle

\begin{abstract}
This paper proposes a system for automatic surface volume monitoring from time series of SkySat pushframe imagery. A specific challenge of building and comparing large 3D models from SkySat data is to correct inconsistencies between the camera models associated to the multiple views  that are necessary to cover the area at a given time, where these camera models are represented as Rational Polynomial Cameras (RPCs). We address the problem by proposing a date-wise RPC refinement, able to handle dynamic areas covered by sets of partially overlapping views. The cameras are refined by means of a rotation that compensates for errors due to inaccurate knowledge of the satellite attitude. The refined RPCs are then used to reconstruct multiple consistent Digital Surface Models (DSMs) from different stereo pairs at each date. RPC refinement strengthens the consistency between the DSMs of each date, which is extremely beneficial to accurately measure volumes in the 3D surface models. The system is tested in a real case scenario, to monitor large coal stockpiles. Our volume estimates are validated with measurements collected on site in the same period of time.
\end{abstract}
\section{Introduction}
\label{sec:intro}

Stockpile measurement is a task of major importance in a wide variety of industrial activities involving the storage, treatment and transport of bulk materials (e.g. mining \cite{raeva2016volume}, landfill management \cite{tucci2019monitoring}).

\begin{figure}[t]
\centering
 \includegraphics[width=0.49\textwidth]{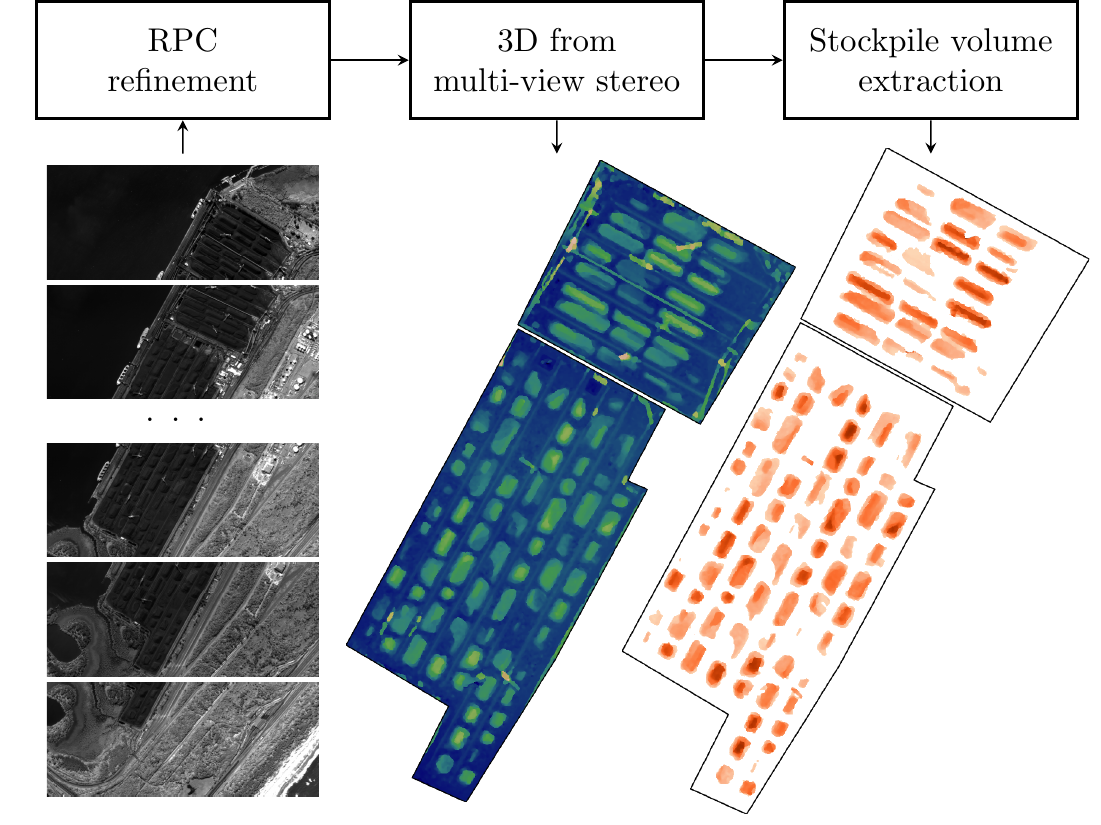}
 \caption[]{Diagram of our surface volume monitoring approach. The RPCs of a time series of SkySat acquisitions are refined and used to compute a high-quality surface model for each date, where volume is measured.}
\label{fig:presented_pipeline}
\end{figure}

Large scale areas represent a particularly adverse scenario to measure vast piles of possibly irregular shapes. Primary methods, such as human-led topographic inspections, can pose a safety risk and are ineffective in terms of time and accuracy. This has raised the interest on technology-aided surveys employing ranging sensors and photogrammetry. Lidar scans and imagery from Unmanned Aerial Vehicles (UAVs), planes or robots, have proven to be highly accurate and have gained great popularity in recent years \cite{raeva2016volume, tucci2019monitoring, he2019ground, zhang2019lidar}.

However, aerial or ground based solutions are often costly or not viable in restricted areas, which may prevent frequent updates. Recurrent satellite imagery is hence emerging as an alternative, as it allows to classify, segment and reconstruct large areas without on-site actions \cite{schmidt2015satellite, salim2017monitoring, defranchis2014automatic, dautume2020stockpile}. \mbox{Today's} abundance of Earth observation satellite images, notably due to the breakthrough of \textit{SmallSats}, explains this growing trend. SmallSat constellations have many low-cost satellites flying at low-altitude orbits, that produce high-resolution images with short revisit times, e.g. PlanetScope, SkySat~\cite{curzi2020large}.

Unlike traditional satellite imagery, SkySat products are delivered as mosaics of images, denoted as \emph{scenes}, with a small geographic footprint. To this end, the pushframe acquisition systems ensure a certain overlap between consecutive scenes \cite{planet2019specs}. Each scene is provided with a Rational Polynomial Camera (RPC) model, but the different RPCs associated to a mosaic typically exhibit non-negligible bias. Inconsistencies between camera models lead to systematic errors in depth estimation from image correspondences, hindering multi-view 3D reconstruction. RPC correction methods are indispensable to exploit such fragmented data, which is in contrast to the case of large-footprint satellite images \cite{mari2019bundle}. In addition, highly dynamic areas like open-air storage facilities pose an extra challenge for RPC refinement: images acquired at different dates should be handled without assuming a static scene over time.

Motivated by the previous observations, we present an automatic system to monitor stockpiles or large irregular volumes, in general, from time series of SkySat scenes. The method can be applied to stereo and tristereo acquisitions \cite{planet2019specs}. Our contributions are:

\begin{itemize}
    \item [-] A generic date-wise RPC refinement, independent of satellite specificities and able to handle areas that change over time.

    \item [-] A volume tracking strategy based on a time series of high-quality surface models obtained by multi-view 3D reconstruction using the refined RPCs.

    \item [-] A performance validation based on data acquired on site.
\end{itemize}

\section{Related work}
\label{sec:related_work}

\subsection{Stockpile volume monitoring}

Most image based approaches to compute stockpile volume perform a 3D reconstruction of the area based on the dense matching of multiple views \cite{raeva2016volume, tucci2019monitoring, he2019ground, schmidt2015satellite}. Single image methods using site-specific heuristics or shape from shading have also been explored \cite{dautume2020stockpile}. 

Once the 3D geometry of the scene is computed, stockpile volume can be measured in different ways. Cross-section methods model piles as big fairly regular solids, while horizontal section methods divide them into layers following contour lines \cite{tucci2019monitoring}. For finer estimations on irregular shapes, it is common to discretize the scene into small elementary 3D volumes, i.e. voxels, tetrahedrons, trigonal prisms \cite{zhu2018accurate, zhang2019lidar}; or into Digital Surface Models (DSMs), i.e. a 2D grid where each cell is assigned an altitude value \cite{raeva2016volume, arango2015comparison}. The boundaries of the piles are usually obtained by subtraction of a bare terrain model, possibly combined with a segmentation step  \cite{he2019ground, tucci2019monitoring}.

\subsection{RPC model refinement}

The RPC model is a generic camera model, independent from specific physical properties, widely used to describe satellite optical sensors. The RPC of a satellite image relates 3D space coordinates (latitude, longitude, height) to 2D image coordinates (pixel row and column). The 3D to 2D mapping and its inverse are known, respectively, as the \textit{projection} and \textit{localization} functions. 

In practice, commercial satellite imagery RPCs contain inaccuracies caused by internal measurement errors of the complex physical system they encode, which need to be refined \cite{grodecki2003block}. RPC refinement strategies typically rely on a set of tie-points, whose projection across the input views has to coincide. Bundle adjustment formulations \cite{triggs1999bundle} are a well known solution, which seeks to minimize the reprojection error of the tie-points by optimizing a set of parameters that modify the original camera models. \cite{grodecki2003block} proved that composing each RPC projection with a 2D offset suffices to adjust satellite images covering lengths of 50 km or less, as this compensates the main source of errors, i.e. inaccuracies in sensor attitude. An equivalent solution to remove such errors is to apply a correction rotation around each camera center previous to the RPC projection \cite{mari2019bundle}. Functions of polynomial form, defined in object or image space, depending on whether they are applied before or after the input RPC mappings, can be used to additionally handle higher order error sources (e.g. time-dependent drift, lens distortion) \cite{grodecki2003block}.
\section{Methodology}
\label{sec:methodology}

\subsection{Bundle adjustment to correct camera orientation}
\label{sec:bundle_adjustment}

Since we do not have access to Ground Control Points (GCPs) of the study area, i.e. points whose object and image coordinates are known in advance, we perform a relative RPC correction based on tie-points derived from feature correspondences. The correction is done in a date-wise manner, solving an independent bundle adjustment problem for the group of cameras of each acquisition date. This ensures that the geometry seen by the cameras is coherent, which is not guaranteed if multiple dates are treated at once. Thus, feature mismatches are minimized and the accuracy of tie-points increases.

Note that a relative correction is sufficient to register the RPCs in a common frame of reference and use them for 3D reconstruction, but the absolute location of the scene remains subject to the geolocation accuracy of the input models. To avoid large drifts in object space, we fix a reference camera for each date, which is not refined.

We refer to the list of image coordinates containing the location of a tie-point across multiple images as a \textit{feature track}. SIFT keypoints~\cite{lowe2004distinctive} are extracted for each image and matched to the keypoints of overlapping views with sufficient baseline, using a ratio test of 0.6 and a RANSAC Fundamental matrix geometric filtering \cite{hartley2003multiple}. The union-find algorithm from~\cite{moulon2012unordered} is used to extend stereo correspondences to unordered feature tracks of arbitrary length.

The object space coordinates of the tie-point associated to each feature track are initialized by triangulating all the stereo matches in the track with the input RPC models, as in~\cite{defranchis2014automatic}, and taking the mean of the 3D locations. All tie-point coordinates are afterwards optimized simultaneously with the correction parameters of each camera.

Our method refines the RPC models by composing them with a preceding rotation around the camera center. The bundle adjustment problem is therefore expressed as
\begin{equation}
     \min_{R_m, X_{k}} \sum_{k=1}^{K}\sum_{m=1}^{M} \| P_m  (R_m (X_k - C_m) + C_m) - x_{mk} \|^2,
     \label{eq:reprojectionerror}
\end{equation}
where Equation~\ref{eq:reprojectionerror} represents the reprojection error of the setting. $X_{k}$ denotes the $k$-th tie-point, $x_{mk}$ its observation on the $m$-th image, and $P_m  (R_m (X_k - C_m) + C_m)$ its reprojection given by the $m$-th RPC projection $P_m$ and the correction rotation $R_m$ around the center $C_m$. The camera center $C_m$ is derived by regressing a projective model from each  RPC model. 

$R_m$ is initialized as the identity matrix and used in the bundle adjustment procedure using the Euler angles representation, which entails 3 variables to be optimized  per camera, i.e.
\begin{align}
\label{eq:eulerangles}
R_m &=  
\scalemath{0.70}{
 \begin{pmatrix} 
 1 & 0 & 0 \\
 0 & \cos{\phi} & -\sin{\phi} \\
 0 & \sin{\phi} & \cos{\phi}
 \end{pmatrix}
 \begin{pmatrix} 
 \cos{\theta} & 0 & \sin{\theta} \\
 0 & 1 & 0 \\
 -\sin{\theta} & 0 & \cos{\theta}
 \end{pmatrix}
 \begin{pmatrix} 
 \cos{\alpha} & -\sin{\alpha} & 0 \\
 \sin{\alpha} & \cos{\alpha} & 0 \\
 0 & 0 & 1
 \end{pmatrix}},
\end{align}
where $\phi$, $\theta$, $\alpha$ are the 3 Euler angles associated to the $m$-th camera.

\begin{figure*}[t]
\centering
\includegraphics[width=\textwidth]{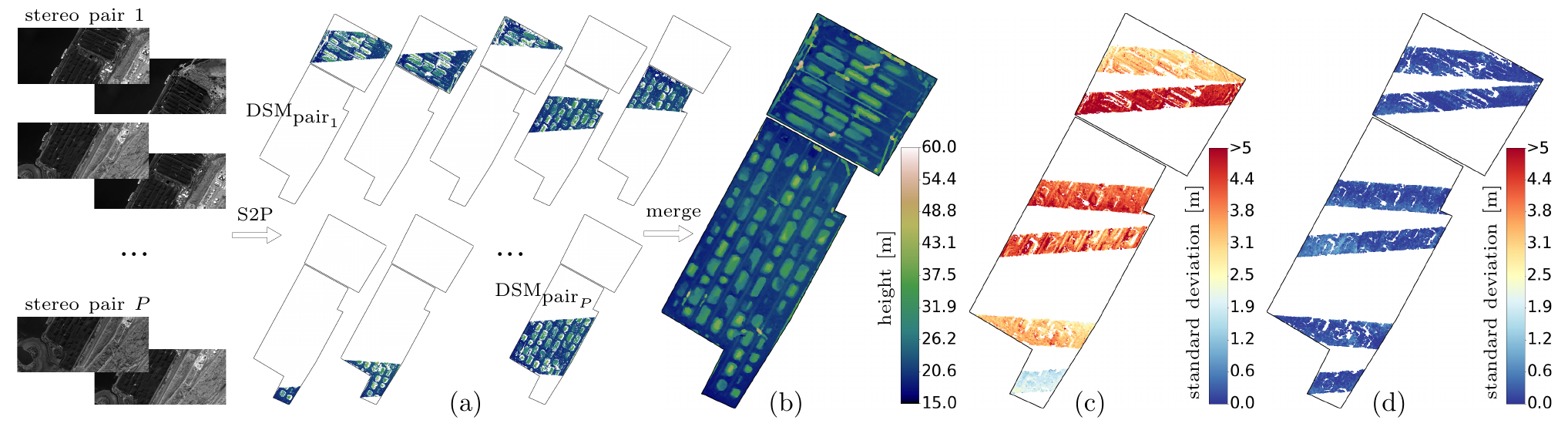}
  \caption{Illustration of the multi-view stereo process run to reconstruct the study area at each date. (a) Several DSMs of the different parts of the area are computed independently from different stereo pairs. (b) Thanks to RPC correction, the stereo DSMs are accurately registered and can be merged directly by taking the average height at each 2D cell. (c) and (d) show, respectively, the standard deviation, in meters, between height values of overlapping stereo DSMs with and without RPC refinement (the DSMs  overlap only on the shown strips).}
    \label{fig:3d_rec}
\end{figure*}

\begin{figure}[t]
  \centering
  \includegraphics[width=0.49\textwidth]{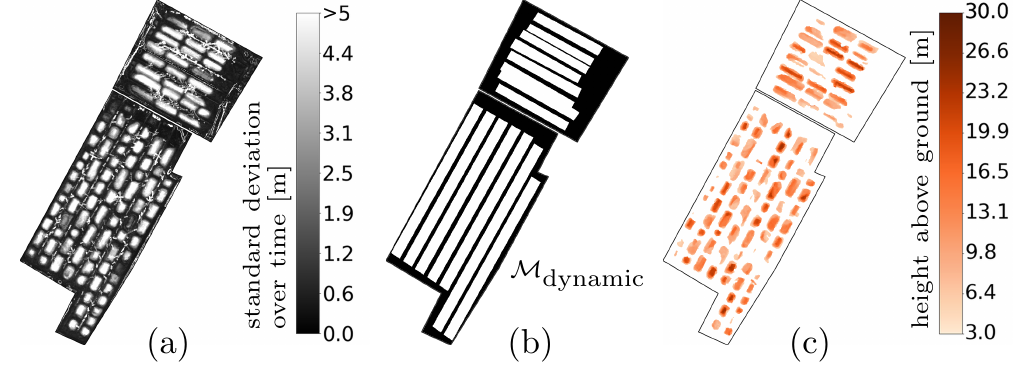}
  \caption{(a) Point-wise standard deviation of height values along the DSM time series. (b) Mask of dynamic parts. (c) Example of nDSM.}
    \label{fig:dynamic_mask_nDSM}
\end{figure}

\begin{figure*}[t]
  \centering
  \includegraphics[width=\textwidth]{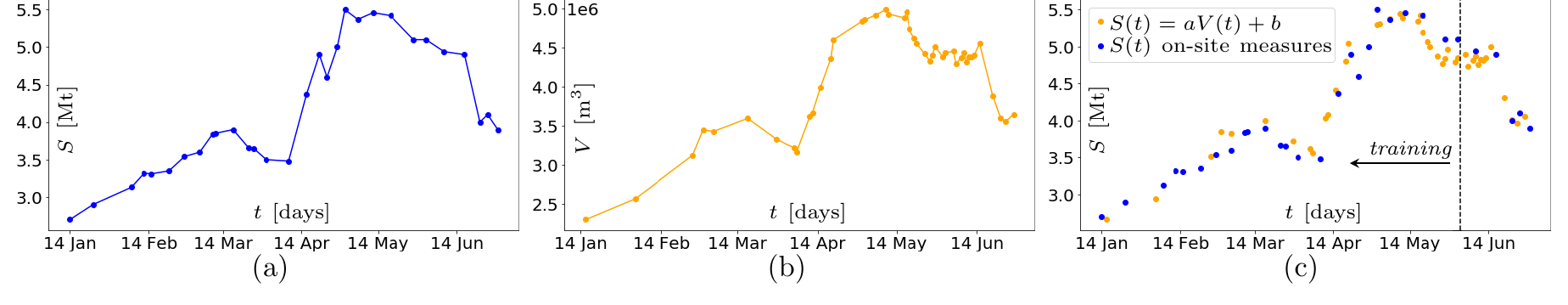}
  \caption{(a) On-site coal stock measurements, $S$, in mega tonnes (Mt). Both in this plot and (b) the line linking the set of discrete measures, represented as dots, is linearly interpolated. (b) Stockpile volume, $V$, as obtained from the time series of photogrammetric DSMs derived from satellite imagery. (c) Linear least-squares regression to predict coal mega tonnes as a function of stockpile volume estimates.}
    \label{fig:results}
\end{figure*}

\subsection{Multi-view stereo reconstruction}
\label{sec:mvs3d}

After RPC refinement, the open-source satellite stereo pipeline S2P\footnote{\url{https://github.com/cmla/s2p}}~\cite{defranchis2014automatic} is run with the corrected camera models. For each date S2P is used to reconstruct DSMs of the parts of the study area seen in all pairs of overlapping scenes that meet certain criteria (Fig.~\ref{fig:3d_rec}a): 
the convergence angle of a pair of scenes must be more than 5 degrees and less than 35 degrees; and the bounding box of the triple intersection between the two scenes footprints and the study area must have both its dimensions larger than 200 meters.

\noindent Reconstructions from different pairs are natively registered in object space as a result of RPC correction. This allows to easily merge them into a denser and highly accurate model of the entire area by taking the average height value at each DSM cell (Fig.~\ref{fig:3d_rec}b). Small holes in the DSMs, due to occlusions between views or lack of texture, are filled by a $5 \times 5$ median filter followed by cubic interpolation. The Ground Sample Distance (GSD) is set to 1 m for all DSMs.

Similarly to~\cite{facciolo2017automatic}, the DSMs of different dates are aligned by a 3D translation that maximizes the Normalized Cross Correlation between their geometry. Even if the geometry may change between different dates, this alignment serves to minimize the standard deviation of DSM heights over time, which is later exploited to determine a coarse mask of the dynamic parts of the study area (Fig.~\ref{fig:dynamic_mask_nDSM}).

Fig.~\ref{fig:3d_rec}c and \ref{fig:3d_rec}d highlight the impact of RPC refinement: the point-wise deviation between height values of different stereo DSMs is of the order of a few meters without the bundle adjustment from Section~\ref{sec:bundle_adjustment}, while it is reduced to tens of centimeters afterwards. 

\subsection{Volume estimation}
\label{sec:volume_monitoring}

Time series of aligned DSMs from remote sensing can be effectively employed to measure and track volumes over an area of interest. As a concrete example of this, we propose an automatic system to monitor stockpiles volume in a real case scenario (Section~\ref{sec:data}).

For each DSM, a Digital Terrain Model (DTM) is subtracted to consider only the heights above ground. Since our study area lies on a flat terrain, we model the DTM as a plane with  height equal to the 25th percentile of the DSM heights. Without loss of generality, cloth simulation methods can be used to model non-flat DTMs \cite{zhang2016easy}.

Additionally, we determine a site-specific mask $\mathcal{M}_{\text{dynamic}}$ delimiting the dynamic parts of the area (Fig.~\ref{fig:dynamic_mask_nDSM}b). The labeling of $\mathcal{M}_{\text{dynamic}}$ is based on the point-wise standard deviation of height values over time, across the different DSMs of the area (Fig.~\ref{fig:dynamic_mask_nDSM}a). 

\noindent The normalized DSM (nDSM) containing the heights above ground in areas where changes are expected can be expressed as 
\begin{equation}
    \text{nDSM($t$)} = \mathcal{M}_{\text{dynamic}}(\text{DSM($t$) - DTM($t$)})
\end{equation}
where $t$ is the acquisition date of the time series. 

Furthermore, only values in nDSM($t$) between 3 and 30 meters are kept (Fig.~\ref{fig:dynamic_mask_nDSM}c). Values outside this range are likely to be due to noise, surface roughness or machinery and cranes working in the area. Note that both height thresholding and the use of $\mathcal{M}_{\text{dynamic}}$ are site and task specific post-processing steps aimed to reduce noise and ensure that only height values associated to stockpiles are left in nDSM($t$). In the absence of any prior knowledge of the facilities or target volumes, these post-processing steps may be omitted at the expense of a small loss of accuracy.

Finally, the volume of the stockpiles left in nDSM($t$) is computed in cubic meters as the addition of all individual cell volumes:
\begin{equation}
    V(\text{nDSM($t$)}) = \sum_{i} l_i w_i h_i 
\end{equation}
where $l_i, w_i$, $h_i$ represent the length, width and height of the $i$-th cell. We use squared cell DSMs, i.e. $l_i = w_i = $ GSD = 1~m.

\section{Experiments}
\label{sec:experiments}

\subsection{Data}
\label{sec:data}

We tested our method on a time series of SkySat panchromatic L1B scenes covering the Richards Bay Coal Terminal (RBCT) in South Africa, which has an open-air storage area of {\footnotesize $\sim$}1.6 km$^2$. The RBCT is one of the world's leading coal export terminals. Tonnes of coal stockpiles are managed 24 hours a day to be shipped overseas.

The time series comprises 43 acquisition dates, distributed non-uniformly between January and July 2020. Distance between consecutive dates oscillates between 1 and 20 days, falling below 1 week in most cases. For each date there are 6 to 10 scenes, captured by the same sensor among the 3 sensors of the SkySats, with a difference of a few seconds. All images are free from clouds in the study area.

SkySat L1B scenes have a nadir GSD of {\footnotesize $\sim$}0.72 m and a total size of $1349\times3199$ pixels. Each scene is delivered with a RPC camera model. The absolute geolocation accuracy of the provided RPC models is of {\footnotesize $\sim$}30 m, with SkySats orbiting at an altitude of {\footnotesize $\sim$}500 km~\cite{planet2019specs}.

\subsection{Validation using on-site stock weight measurements}

The volume $V$ of coal stockpiles for each date $t$ of the sequence was computed using the system described in Section~\ref{sec:methodology} (Fig.~\ref{fig:results}b). Coal stock weight measurements, $S$, as collected by agents on site during the same period of time, are shown in  Fig.~\ref{fig:results}a in mega tonnes (Mt). The correlation between both sets is recognizable to the naked eye.

To assess the performance of our system, we propose a simple approach to predict coal weight from stockpiles volume, so that the available measurements can be compared in equivalent units. The conversion is not straight-forward, as multiple date-dependent factors may be involved (e.g. humidity factors, non-uniform coal types or pile densities). To this end, we linearly interpolate the two sets of measurements and apply a least-squares regression to fit two coefficients $a$ and $b$ satisfying $S(t) = aV(t) + b$. The result is shown in Fig.~\ref{fig:results}c. We obtain $a=1.02$, $b=0.3$, where $a$ can be interpreted as the best-fitting bulk density and $b$ as a ground offset. Observe that 85\% of the interpolated samples were used to fit $a$ and $b$ (the \textit{training} set), but the strong correlation extends to the rest of dates, stressing the robustness of the method.

It remains difficult to quantify the exact accuracy of the system without measurements that match exactly in time, specially since shipments or stock arrivals can occur within a few hours. In general, Fig.~\ref{fig:results}c seems to indicate that the differences between our weight estimates derived from remote sensing and on-site measurements from neighbor dates are typically < 0.3 Mt. Discrepancies can be probably explained by noisy or interpolated data in the photogrammetric DSMs employed for volume estimation, specific stockpile properties or a strong abrupt activity in the area.

\section{Conclusion}
\label{sec:conclusions}

An automatic system for surface volume monitoring using recurrent satellite imagery was presented. The system is based on a generic RPC refinement step, independent from satellite specificities, which enables to accurately measure volume from multiple independent DSMs obtained with an open-source satellite stereo pipeline. A time series of SkySat images distributed over \mbox{{\footnotesize $\sim$}6 months} was used to validate the  system in a real case scenario concerning a highly dynamic area of coal stockpiles.

\section*{Acknowledgements}
This work was supported by a grant from Région Île-de-France. It was also partly financed by IDEX Paris-Saclay IDI 2016, ANR-11-IDEX-0003-02, Office  of Naval research grant N00014-17-1-2552, DGA Astrid project  \mbox{\guillemotleft \ filmer la Terre \guillemotright} \ n\textsuperscript{o} ANR-17-ASTR-0013-01, MENRT

\bibliographystyle{IEEEbib}
\bibliography{refs}

\begin{thebibliography}{10}

\bibitem{raeva2016volume}
P.L. Raeva, S.L. Filipova, and D.G. Filipov,
\newblock ``{Volume computation of a stockpile - A study case comparing GPS and
  UAV measurements in an open pit quarry},''
\newblock {\em \mbox{ISPRS}}, vol. 41-B1, 2016.

\bibitem{tucci2019monitoring}
G.~Tucci, A.~Gebbia, A.~Conti, L.~Fiorini, and C.~Lubello,
\newblock ``{Monitoring and computation of the volumes of stockpiles of bulk
  material by means of UAV photogrammetric surveying},''
\newblock {\em Remote Sensing}, vol. 11, no. 12, pp. 1471, 2019.

\bibitem{he2019ground}
H.~He, T.~Chen, H.~Zeng, and S.~Huang,
\newblock ``Ground control point-free unmanned aerial vehicle-based
  photogrammetry for volume estimation of stockpiles carried on barges,''
\newblock {\em Sensors}, vol. 19, no. 16, pp. 3534, 2019.

\bibitem{zhang2019lidar}
W.~Zhang and D.~Yang,
\newblock ``{Lidar-based fast 3D stockpile modeling},''
\newblock in {\em ICICAS}, 2019, pp. 703--707.

\bibitem{schmidt2015satellite}
B.~Schmidt, M.~Malgesini, J.~Turner, and J.~Reinson,
\newblock ``Satellite monitoring of a large tailings storage facility,''
\newblock in {\em Tailings and Mine Waste}, 2015.

\bibitem{salim2017monitoring}
P.M. Salim, M.A. Jais, N.~Sahriman, A.M. Samad, M.A. Abbas, I.~Maarof, and
  N.~Tarmizi,
\newblock ``Monitoring quarry areas using remote sensing techniques,''
\newblock in {\em CSPA}, 2017, pp. 323--328.

\bibitem{defranchis2014automatic}
C.~de~Franchis, E.~Meinhardt-Llopis, J.~Michel, J-M. Morel, and G.~Facciolo,
\newblock ``An automatic and modular stereo pipeline for pushbroom images,''
\newblock {\em ISPRS}, vol. 2-3, pp. 49--56, 2014.

\bibitem{dautume2020stockpile}
M.~d'Autume, A.~Perry, J-M. Morel, E.~Meinhardt-Llopis, and G.~Facciolo,
\newblock ``{Stockpile monitoring using linear shape-from-shading on
  PlanetScope imagery},''
\newblock {\em ISPRS}, vol. 5-2, 2020.

\bibitem{curzi2020large}
G.~Curzi, D.~Modenini, and P.~Tortora,
\newblock ``Large constellations of small satellites: A survey of near future
  challenges and missions,''
\newblock {\em Aerospace}, vol. 7, no. 9, pp. 133, 2020.

\bibitem{planet2019specs}
Planet Labs,
\newblock ``Planet imagery product specifications,'' 2019.

\bibitem{mari2019bundle}
R.~Mar{\'\i}, C.~de~Franchis, E.~Meinhardt-Llopis, and G.~Facciolo,
\newblock ``To bundle adjust or not: A comparison of relative geolocation
  correction strategies for satellite multi-view stereo,''
\newblock in {\em ICCV Workshops}, 2019.

\bibitem{zhu2018accurate}
J.~Zhu, J.~Yang, J.~Fan, D.~Ai, Y.~Jiang, H.~Song, and Y.~Wang,
\newblock ``Accurate measurement of granary stockpile volume based on fast
  registration of multi-station scans,''
\newblock {\em Remote Sensing Letters}, vol. 9, no. 6, pp. 569--577, 2018.

\bibitem{arango2015comparison}
C.~Arango and C.A. Morales,
\newblock ``{Comparison between multicopter UAV and total station for
  estimating stockpile volumes},''
\newblock {\em ISPRS}, vol. 40-1, 2015.

\bibitem{grodecki2003block}
J.~Grodecki and G.~Dial,
\newblock ``Block adjustment of high-resolution satellite images described by
  rational polynomials,''
\newblock {\em Photogrammetric Engineering \& Remote Sensing}, vol. 69-1, 2003.

\bibitem{triggs1999bundle}
B.~Triggs, P.F. McLauchlan, R.I. Hartley, and A.W. Fitzgibbon,
\newblock ``{Bundle adjustment — A modern synthesis},''
\newblock in {\em International Workshop on Vision Algorithms}, 1999.

\bibitem{lowe2004distinctive}
D.G. Lowe,
\newblock ``Distinctive image features from scale-invariant keypoints,''
\newblock {\em IJCV}, vol. 60, no. 2, pp. 91--110, 2004.

\bibitem{hartley2003multiple}
R.I. Hartley and A.~Zisserman,
\newblock {\em Multiple view geometry in computer vision},
\newblock Cambridge University Press, 2003.

\bibitem{moulon2012unordered}
P.~Moulon and P.~Monasse,
\newblock ``Unordered feature tracking made fast and easy,''
\newblock in {\em CVMP}, 2012.

\bibitem{facciolo2017automatic}
G.~Facciolo, C.~de~Franchis, and E.~Meinhardt-Llopis,
\newblock ``{Automatic 3D reconstruction from multi-date satellite images},''
\newblock in {\em CVPR Workshops}, 2017.

\bibitem{zhang2016easy}
W.~Zhang, J.~Qi, P.~Wan, H.~Wang, D.~Xie, X.~Wang, and G.~Yan,
\newblock ``{An easy-to-use airborne lidar data filtering method based on cloth
  simulation},''
\newblock {\em Remote Sensing}, vol. 8-6, 2016.

\end{thebibliography}

\end{document}